\documentclass{article}
\usepackage{spconf,amsmath,epsfig}
\usepackage{amsfonts,amssymb}

\title{Super-resolution reconstruction of hyperspectral images via low rank tensor modeling and total variation regularization}
\name{Shiying He$^{1,3}$, Haiwei Zhou$^{1}$, Yao Wang$^{1,3}$, Wenfei Cao$^{2}$, and Zhi Han$^3$ \thanks{This work was supported in part by the  Natural Science Foundation of China under grant numbers 11501440, 61273020 and 61303168. \textit{(Corresponding author: Yao Wang, email: yao.s.wang@gmail.com.)}}}
\address{$^1$School of Mathematics and Statistics, Xi'an Jiaotong University \\
               $^2$School of Mathematics and Information Science, Shaanxi Normal University \\
               $^3$Shenyang Institute of Automation, Chinese Academy of Sciences}
\begin{document}
%
\maketitle
\begin{abstract}
In this paper, we propose a novel approach to hyperspectral image super-resolution by modeling the global spatial-and-spectral correlation and local smoothness properties over
hyperspectral images. Specifically, we utilize the tensor nuclear norm and tensor folded-concave penalty functions to describe the global spatial-and-spectral correlation hidden in hyperspectral images, and 3D total variation (TV) to characterize the local spatial-and-spectral smoothness across all  hyperspectral bands. Then, we develop an efficient algorithm for solving the resulting optimization problem by combing the local linear approximation (LLA) strategy and alternative direction method of multipliers (ADMM).
Experimental results on one hyperspectral image dataset illustrate the merits of the proposed approach .
\end{abstract}
\begin{keywords}
Hyperspectral images, Super-resolution reconstruction, nuclear norm, Folded-concave penalty, 3D total variation.
\end{keywords}
\section{Introduction}
Hyperspectral images (HSIs) are recordings of reflectance of light of some real world scenes or objects including hundreds of spectral bands ranging from ultraviolet to infrared wavelength~\cite{Intr1,Intr2}.
The abundant spectral bands of HSIs provide fine spectral feature differences between various materials of interest and enable many computer vision tasks more successfully achievable.
However, due to the constraints of imaging hardware, signal to noise ratio (SNR) and time constraints, the acquired hyperspectral images unfortunately have low spatial resolution, which cannot give any active help for high precision processing requirements in many fields including mineralogy, manufacturing, medical diagnostics, and surveillance. 
Hence, the task of reconstructing a hyperspectral image of high resolution (HR) from an observed low resolution (LR) hyperspectral image or sequence is a valuable research issue.

The problem of  hyperspectral image super-resolution (HSSR) can be solved by designing various traditional signal processing techniques, including the works \cite{SR3, SR4, TV1}. 
In the recent years, applying prior information of HR auxiliary images into the process of HSSR has been becoming more and more popular~\cite{spatial,Review}. However, such HR images are not always easy to get due to the limitations of remote sensing system. Therefore, super-resolution of single HSI cube has atracted increased interest in many practical scenarios .

In this paper, we consider a single HSI cube as a tensor with three modes (width, height, and band)
and then discover the hidden spatial-and-spectral structures using tensor modelling for enhancing its spatial resolution.
Specifically, the spectral bands of a HSI have strong correlations and each band  if considered as a matrix has relatively strong correlation; this spatial-and-spectral correlation can be modelled
by a low-rank tensor penalty. Additionally, for each voxel, from the spatial viewpoint its intensity seems to
almost equal to those in its neighbourhood, and the same from the spectral viewpoint;
we then describe this local spatial-and-spectral smoothness property using 3D total variation.
As such, the HSSR task resorts to solving an optimization problem, which can be efficiently solved by combing LLA strategy and ADMM. 
\section{HSSR via total variation and low-rank regularizations}
In this section, we first introduce the observation model. Then, we utilize 3D TV to describe 
local smoothness of a hyperspectral image, and adopt a tensor folded-concave penalty  to characterize global correlation of a hyperspectral image.
Finally, a novel regularization model is derived for the HSSR task.
\subsection{Observation model}
The low spatial resolution hyperspectral image can be generated by the following observation model:
$$\mathcal{I}=DS\mathcal{X}+\mathbf{e},$$
where the tensor $\mathcal{I}$ donates the observed LR image, $D$ is a downsampling operator, $S$ is a blurring operator, $\mathcal{X}$ is the HR image to be reconstructed and $\mathbf{e}$ represents the observation noise. Since this is an ill-posed problem, some regularization terms of $\mathcal{X}$ based on prior knowledge, denoted by $\mathfrak{R}(\mathcal{X})$, can be introduced to regularize the solution to refine the solution space: $\hat{\mathcal{X}}=\arg\min_{\mathcal{X}}\{\|DS\mathcal{X}-\mathcal{I}\|^2+\lambda\mathfrak{R}(\mathcal{X})\}$, where $\lambda$ is a scalar parameter to make a trade-off between the fidelity term and the regularization term.
\subsection{3D TV regularization}
Total variation (TV) \cite{TV1} is often used to preserve local spatial consistency in image recovery and suppress image noise. Considering the fact that an HR hyperspecctral image to be reconstructed is treated as a tensor, and its local spatial-and-spectral consistency, or say, smoothness ccharacterized by 3D total variation, which is expressed as 
$TV(\mathcal{X})=\sum_{ijk} |x_{ijk}- x_{ij,k-1}| + |x_{ijk} - x_{i,j-1,k}| + |x_{ijk} - x_{i-1,j,k}|,$ where $x_{ijk}$ is the $(i,j,k)$-th entry of tensor $\mathcal{X}$.
\subsection{Low-rank regularization}
The spatial-and-spectral correlation of a hyperspectral image implies that
each unfolded matrix, if a hyperspectral image represented as a tensor, is low rank.
Hence, following the work~\cite{ten2}, low-rank property of a three-order tensor can be measured by a weighted sum of three ranks:
\begin{equation}
\label{ten_rank}
\text{Rank}(\mathcal{X})=\sum_{i}^{3}\alpha_i \text{Rank}(\mathcal{X}_{(i)}),
\end{equation}
where $\alpha_i\geqslant 0$ and satisfies $\sum_{i}^{3}\alpha_i=1$.
Since the optimization problem with rank constraint (1) is intractable, and matrix nuclear norm is exploited as a tight convex
surrogate of the matrix rank~\cite{FC2} , one can replace the rank function \eqref{ten_rank} with  the following tensor nuclear norm:
\begin{equation}
\label{ten_norm}
\|\mathcal{X}\|_{*}=\sum_{i}^{3}\alpha_i \|\mathcal{X}_{(i)}\|_{*},
\end{equation}
where $\|\mathbf{Z}\|_{*}:=\sum_{k=1}^{\min(m,n)}\sigma_k(\mathbf{Z})$ denotes the nuclear norm of matrix $\mathbf{Z}$ of size $m\times n$, and $\mathcal{X}_{(i)}$ is the $i$-th unfolded matrix of tensor $\mathcal{X}$~\cite{ten2}. 

Although the convex nuclear norm  \eqref{ten_norm} performs well in various tensor recovery problems, studies such as \cite{FC2} have shown that the nuclear norm over-penalizes large singular values, and thus leads to the modeling bias in low rank structure estimation. Folded-concave penalty~\cite{FC3} can be used to remedy this modeling bias, as shown in some works \cite{FC3, FC}. Thus, we shall utilize one of the folded penalties, the minmax concave plus (MCP) penalty, of the form:
\begin{equation}
P_{\lambda}=
   \begin{cases}
   a\lambda^2/2 &\mbox{if $|t|\geqslant a\lambda$}\\
   \lambda|t|-\frac{t^2}{2a} &\mbox{otherwise}.
   \end{cases}
\end{equation}
Following \cite{FC}, the folded-concave norm of a matrix $X$ is defined as $\|X\|_{P_\lambda}:=\sum_{j=1}^{r}P_{\lambda}\big(\sigma_j(X)\big)$\footnote{Note that $\|X\|_{P_{\lambda}}$ is nonconvex with respect to $X$.}, where $\sigma_j(X)$ is the $j$-th singular value of $X$ and $r$ is the rank. As such, the tensor MCP penalty is defined by applying the MCP penalty function to each unfolded matrix $\mathcal{X}_{(i)}$:
\begin{equation}
\label{ten_mcp}
\|\mathcal{X}\|_{P_{\lambda}}=\sum_{i}^{N}\alpha_i\|\mathcal{X}_{(i)}\|_{P_{\lambda}}.
\end{equation}
\subsection{Proposed model}
Based on the previous discussions, we now derive the following regularization model for the HSSR task:
\begin{equation}
\label{ten_rec}
\min_{\mathcal{X}}\|DS\mathcal{X}-\mathcal{I}\|^2_F+\lambda_1TV(\mathcal{X})+\lambda_2\mathcal{L}(\mathcal{X}_{(i)}), 
\end{equation}
where the scalars $\lambda_1$ and $\lambda_2$ are regularization parameters, and $\mathcal{L}(\mathcal{X}_{(i)})$ is the low-rank measure function (1) or (4) for $\mathcal{X}$. 
\section{Optimization Algorithm}
\label{sec:pagestyle}
We first rewrite (5) as the following equivalent form by introducing $N$ auxiliary variable $\{\mathcal{M}_{i}\}_{i=1}^N$:
\begin{equation}
\begin{split}
  &\min_{\mathcal{X},\{\mathcal{M}_i\}_{i=1}^N}~~\|DS\mathcal{X}-\mathcal{I}\|^2_F+\lambda_1TV(\mathcal{X})+\lambda_2\mathcal{L}(\mathcal{M}_i)\\
  &    s.t~~\mathcal{X}_{(i)}=\mathcal{M}_{i(i)},i=1,2,...,N
\end{split}
\end{equation}
Based on ADMM~\cite{ADMM}, the augmented Lagrangian function is written as follows: 
\begin{equation}
\begin{split}
   &L(\mathcal{X},\mathcal{Y}_i,\mathcal{M}_i)=\|DS\mathcal{X}-\mathcal{I}\|^2_F+\lambda_1TV(\mathcal{X})\\
   &+\sum_{i=1}^{N}\lambda_2\mathcal{L}(\mathcal{M}_{i(i)})+\sum^{N}_{i=1}\frac{\rho}{2}\|\mathcal{M}_{i(i)}-\mathcal{X}_{(i)}+\frac{\mathcal{Y}_{i(i)}}{\rho}\|^2_F, 
\end{split}
\end{equation}
where $\{\mathcal{Y}_i\}_{i=1}^N$ are Lagrangian parameters. We shall break (7) into three subproblems and iteratively update each variable through fixing the other ones. Let $k$ denotes the $k$th iteration step:

\textbf{Subproblem 1: }
\begin{equation}
\begin{split}
    & \mathcal{X}^{(k+1)}=arg\min_\mathcal{X}~~\|DS\mathcal{X}-\mathcal{I}\|^2_F+\lambda_1TV(\mathcal{X}) \\
    &+\sum^{N}_{i=1}\frac{\rho}{2}\|\mathcal{M}_{i(i)}^k-\mathcal{X}_{(i)}+\frac{\mathcal{Y}_{i(i)}^k}{\rho}\|^2_F
\end{split}
\end{equation}
The well-known gradient  method can be easily applied to solve this subproblem.

\textbf{Subproblem 2: }
\begin{equation}
\begin{split}
    & \{\mathcal{M}_i^{(k+1)}\}_{i=1}^N= arg\min_{\{\mathcal{M}_i\}_{i=1}^N}\sum_{i=1}^{N}\lambda_2\mathcal{L}(\mathcal{M}_{i(i)})\\
    &+\sum^{N}_{i=1}\frac{\rho}{2}\|\mathcal{M}_{i(i)}-\mathcal{X}^k_{(i)}+\frac{\mathcal{Y}_{i(i)}^k}{\rho}\|^2_F
\end{split}
\end{equation}
The solution of this subproblem depends on the choice of the low rank term $\mathcal{L}(\mathcal{X})$. We first consider the case of nuclear norm, i.e., 
\begin{equation}
\sum_{i=1}^{N}\lambda_2\alpha_i\|\mathcal{M}_{i(i)}\|_*+\sum^{N}_{i=1}\frac{\rho}{2}\|\mathcal{M}_{i(i)}-\mathcal{X}^k_{(i)}+\frac{\mathcal{Y}_{i(i)}^k}{\rho}\|^2_F
\end{equation}
According to~\cite{ten2}, its close-form solution is expressed as
\begin{equation}
\mathcal{M}_i=fold_i[S_{\lambda_2\alpha_i/\rho}(\mathcal{X}_{(i)}^{(k+1)})-\mathcal{Y}_{i(i)}^{(k)}]
\end{equation}
For a given matrix $X$, the singular value shrinkage operator $S_{\tau}(X)$ is defined by $S_{\tau}(X):=U_XD_{\tau}(\Sigma_X)V_X^T$, where $X=U_X\sigma_XV_X^T$ is the singular value decomposition of $X$ and $[D_{\tau}(A)]_{ij}=sgn(A_{ij})(|A_{ij}|-\tau)_+$.

While for the MCP case, we adopt the same idea of~\cite{FC3, FC} to solve the resulting nonconvex problem. More precisely, we use the local linear approximation (LLA) algorithm to transform the MCP penalization problem into a series of weighted nuclear norm penalization problem. Then the resulting optimization problem can be solved as well. More precisely, the \textbf{subproblem 2} can be written as
\begin{equation}
\begin{split}
    & \{\mathcal{M}_i^{(k+1)}\}_{i=1}^N= arg\min_{\{\mathcal{M}_i\}_{i=1}^N}\sum_{i=1}^{N}\lambda_2\alpha_iQ_{P_\lambda}(\sigma(\mathcal{M}_{i(i)})|\sigma(\mathcal{X}^{k}))\\
    &+\sum^{N}_{i=1}\frac{\rho}{2}\|\mathcal{M}_{i(i)}-\mathcal{X}^k_{(i)}+\frac{\mathcal{Y}_{i(i)}^k}{\rho}\|^2_F, 
\end{split}
\end{equation}
where $Q_{P_\lambda}(\sigma(X|X_{(i)}^{k}))$ is the locally linear approximation of $\|X\|_{P_\lambda}$ when $X^{k}$ is given. Then the solution of this optimization problem is $\mathcal{M}_{i(i)}=S_{\alpha_i/\rho,W_i}(\mathcal{X}_{(i)}-\frac{\mathcal{Y}_{i(i)}}{\rho})$ and the weight matrix $W_i$ is given by $W_i=Diag((\lambda-(\sigma(\mathcal{X}_{(i)})/a))_+)$ for some fixed $a>1$.

\textbf{Subproblem 3: }
\begin{equation}
  \mathcal{Y}_i^{(k+1)}=\mathcal{Y}_i^{(k)}+\rho(\mathcal{M}_i^{(k+1)}-\mathcal{X}^{(k+1)}), 
\end{equation}
where $\rho$ is a parameter associated with convergence rate with fixed value, i.e., 1.05. 
\section{Experimental study}

We now test the proposed method on a HSI dataset. The reference image without noisy bands is a $256\times256\times146$ hyperspectral image acquired over Moffett field, CA, in 1994 (AVIRIS). The blurring kernel is Gaussian kernel and the LR image is generated by downsampling the original HR image with a factor of 2, i.e.,  the LR image is of size $128\times128\times146$.

We compare our method with three other popular methods, including
the bicubic method described in \cite{Bi}, NARM proposed in \cite{Na} and Sparse Representation method by Yang et al.~\cite{yang}. The reconstructed results of the test HSI for a specific band 100 are shown in Fig.1.
\begin{figure*}[htb]
\centering
\includegraphics[width=500pt]{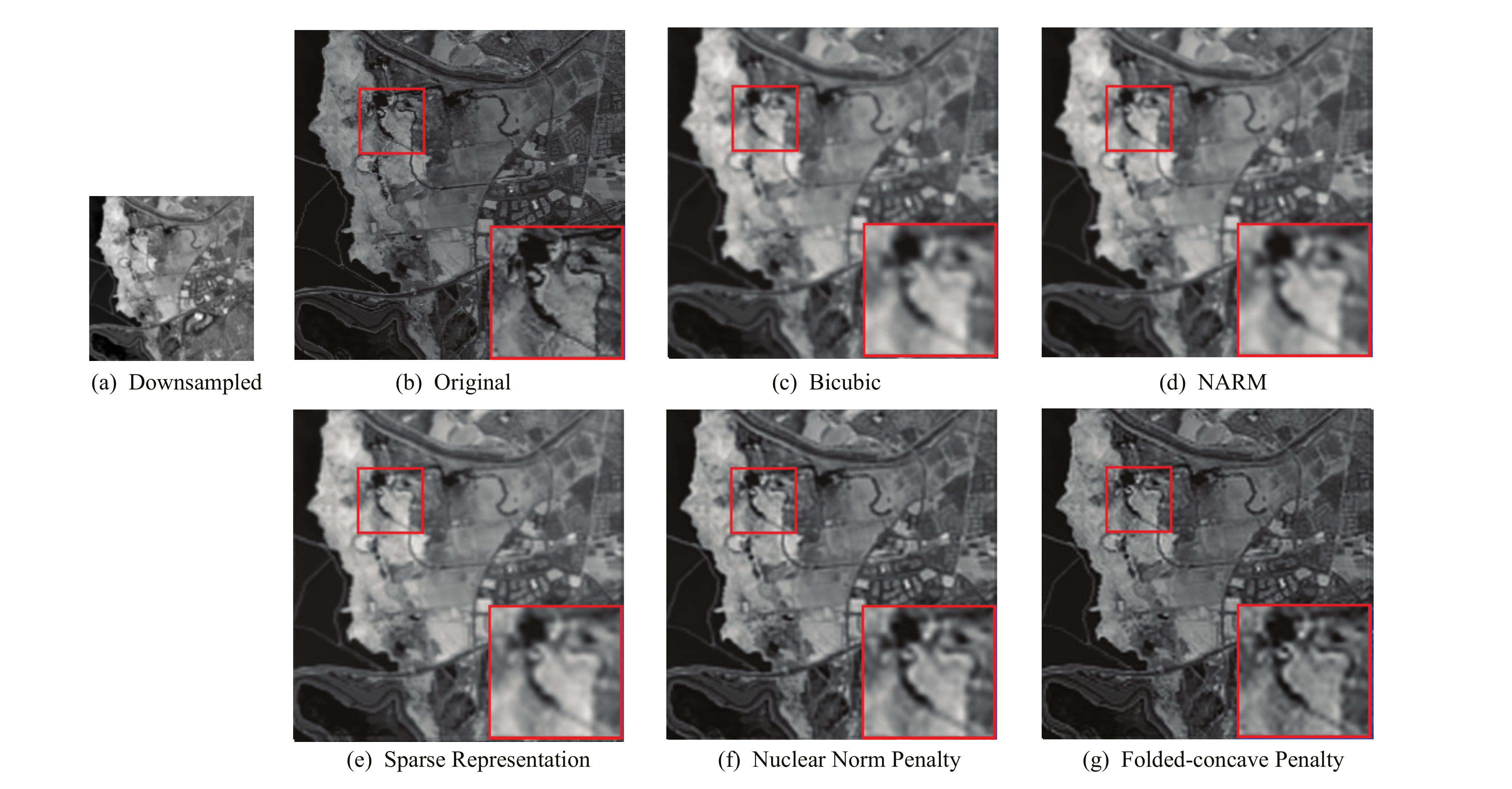}
\caption{Visual comparison of different super-resolution reconstruction methods.}
\end{figure*}
One can observe  that the Bicubic
interpolation blurs the image and the high-frequency spatial details are lost. The other methods provide better reconstruction visual effects. 
Additionally, our proposed method shown in Fig.1(e) and (f) outperforms the other ones. It is also interesting to note that the folded-concave penalization, i.e.,  the MCP,  outperforms other competing methods.

To further evaluate the quality of the proposed reconstruction strategy, several image quality measures have been employed, including peak-signal to noise ratio (PSNR), spectral angle mapper (SAM), and relative dimensionless global error in synthesis (ERGAS). It is known that the larger the PSNR, the better the image quality is; the lower the SAM and ERGAS value are, the smaller spectral distortion.

%
%
\begin{table}[h]
\caption{Quantitative Measures for Different SRR Methods}
\label{table:para}
\centering
\begin{tabular}{|c||c||c||c|}
\hline
Quantitative Measures              &PSNR     &SAM      &ERGAS   \\
\hline
Bicubic                          &33.0236  &0.1248   &126.0507\\
\hline
NRAM                               &33.1197  &0.1297   &124.3686  \\
\hline
Sparse Representation              &35.7409  &0.1651   &117.4637\\
\hline
Nuclear Norm Penalty               &\textit{36.9567 } &\textit{0.0843}   &\textit{95.0166} \\
\hline
MCP Penalty             &\textbf{37.8732}  &\textbf{0.0720}   &\textbf{88.5562}\\
\hline
\end{tabular}
\end{table}
It can be seen from Table 1 that the proposed method with nuclear norm and folded-concave penalties outperforms other competing ones. Again, the MCP penalization provides best reconstruction results, which illustrates the advantage of folded-concave penalty over convex nuclear norm penalty. 
\section{Conclusion}
In this paper, we propose a novel method for hyperspectral  image super-resolution  by tensor
structural modelling. The proposed method considers the global correlation and local smoothness of a hyperspectral image by combining low-rank and total variation regularizations imposed on a tensor. Experimental results reveal that the proposed methods outperform other compared methods, and especially folded concave penalization is superior over the nuclear norm penalization for the HSSR task.

\bibliographystyle{IEEEbib}
\bibliography{igarss}

\end{document}